\documentclass[10pt,onecolumn,letterpaper]{article}

\usepackage{cvpr}
\usepackage{times}
\usepackage{epsfig}
\usepackage{graphicx}
\usepackage{amsmath}
\usepackage{amssymb}

\usepackage{epstopdf}
\usepackage{algorithm}
\usepackage{algorithmic}
\usepackage{subfigure}
\usepackage{booktabs}
\usepackage{bm}
\usepackage{url}

\graphicspath{{./figure/}}

\newcommand{\midfigurewidth}{50}

\usepackage[pagebackref=true,breaklinks=true,letterpaper=true,colorlinks,bookmarks=false]{hyperref}

\cvprfinalcopy 

\def\cvprPaperID{1139} 

\ifcvprfinal\fi
\begin{document}

\title{Exploiting Structure Sparsity for Covariance-based Visual Representation}

\author{Jianjia Zhang\\
{\tt\small jz163@uowmail.edu.au}
\and
Lei Wang\\
{\tt\small leiw@uow.edu.au}
\and
Luping Zhou\\
{\tt\small lupingz@uow.edu.au}
\and
Wanqing Li\\
{\tt\small wanqing@uow.edu.au}
}
\maketitle
{School of Computing and Information Technology. University of Wollongong, Wollongong, 2522, Australia.}
~\\
~\\
\begin{abstract}
The past few years have witnessed increasing research interest on covariance-based feature representation. A variety of methods have been proposed to boost its efficacy, with some recent ones resorting to nonlinear kernel technique. Noting that the essence of this feature representation is to characterise the underlying structure of visual features, this paper argues that an equally, if not more, important approach to boosting its efficacy shall be to improve the quality of this characterisation. Following this idea, we propose to exploit the structure sparsity of visual features in skeletal human action recognition, and compute sparse inverse covariance estimate (SICE) as feature representation. We discuss the advantage of this new representation on dealing with small sample, high dimensionality, and modelling capability. Furthermore, utilising the monotonicity property of SICE, we efficiently generate a hierarchy of SICE matrices to characterise the structure of visual features at different sparsity levels, and two discriminative learning algorithms are then developed to adaptively integrate them to perform recognition. As demonstrated by extensive experiments, the proposed representation leads to significantly improved recognition performance over the state-of-the-art comparable methods. In particular, as a method fully based on linear technique, it is comparable or even better than those employing nonlinear kernel technique. This result well demonstrates the value of exploiting structure sparsity for covariance-based feature representation.   

\end{abstract}

\section{Introduction}
Covariance-based feature representation (Cov-RP in short) uses the covariance matrix of a predefined visual feature vector to represent an image region, a whole image, a set of images, or a sequence of video frames. It has been applied to various vision tasks including object detection and recognition~\cite{DBLP:conf/eccv/TuzelPM06}, action recognition~\cite{DBLP:conf/ijcai/HusseinTGE13,harandi2014manifold}, image set classification~\cite{DBLP:conf/cvpr/WangGDD12}, to name a few. Through these applications, the Cov-RP has gradually developed from a local region descriptor to a more generic visual representation~\cite{kernelrepiccv}. This trend of development is also reflected in the improvements proposed for this representation, from fast computation of covariance region descriptor~\cite{DBLP:conf/eccv/TuzelPM06}, through developing novel measures and algorithms~\cite{quang2014log}, to the recent use of nonlinear kernel techniques to  characterise or even replace covariance matrix~\cite{harandi2014bregman,kernelrepiccv}.   


Some visual recognition tasks face the issues of small sample size and high feature dimensionality. A typical example is skeletal human action recognition since usually a high dimensional feature vector is required to describe each video frame while the number of frames for one action instance is limited. This makes the estimate of covariance matrix unstable or even singular, significantly affecting the effectiveness of Cov-RP. A number of remedies have been proposed to address this fundamental issue, including appending scaled identity matrix~\cite{DBLP:conf/cvpr/WangGDD12}, 
or using kernel matrix instead~\cite{kernelrepiccv}. They have effectively improved the performance of Cov-RP in various recognition tasks. 

In this paper, we note that Cov-RP essentially aims to characterise the underlying structure of visual features. However, the existing remedies have not yet paid sufficient attention to the appropriateness of sample-based covariance matrix for such a goal, and still treated the available samples as the only source of information. These make Cov-RP hard to fit the complexity of the tasks in recent applications.
\begin{figure}[!t]
\label{fig:SICEch}
\begin{center}
\begin{tabular}{ccc}
\hspace{10 mm}
{\includegraphics[height = 35 mm]{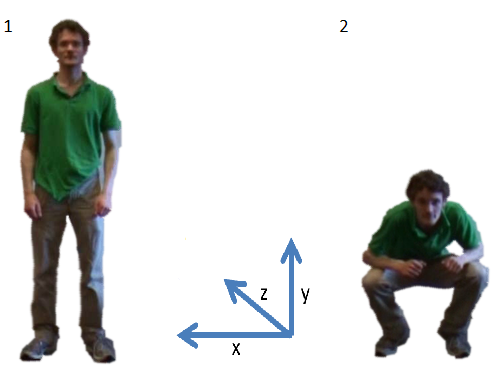}}&
{\includegraphics[width = \midfigurewidth mm]{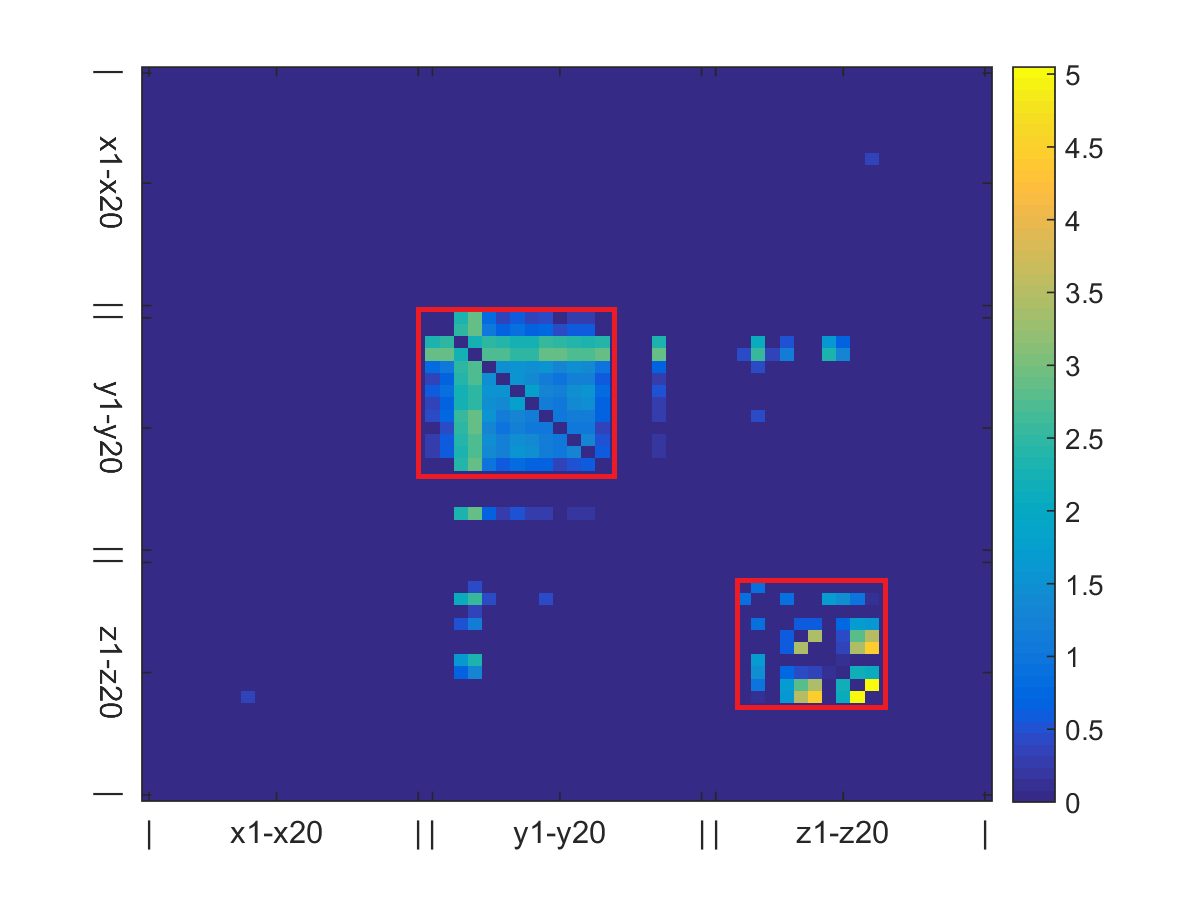}}\\
 (a) ``Crouch or hide'' action \\~~~~~~from MSRC-12 data set. &(b) Proposed SICE-RP\\
\end{tabular}
\end{center}
\caption{Visualisation of SICE representation (SICE-RP) (b) for a ``Crouch or hide'' action (a) from MSRC-12 data set.  SICE-RP responds highly inside the two red boxes, corresponding to the interactions between the $y$-coordinates and the interactions between the $z$-coordinates of the upper body, respectively. It responds lowly to the interactions of x-coordinates. These patterns are well consistent with the fact that the joints to squat down mainly move along $y$- and $z$-axis while staying still along $x$-axis.}
\label{fig:action2}
\end{figure}
To address the above issues, this paper argues that an equally, if not more, important approach to boosting Cov-RP shall focus on better characterising the underlying structure of visual features. In particular, for the task of skeleton-based human action recognition, prior knowledge on feature structure is readily to be used~\cite{lehrmann2013non}, and the most common one may be structure sparsity~\cite{huang2011learning}. Inspired by this observation, this work aims to investigate the effectiveness of exploiting structure sparsity for Cov-RP specifically in skeleton-based human action recognition. To conveniently accommodate this prior knowledge, we migrate from covariance matrix to its inverse, and take advantage of the sparse inverse covariance estimation (SICE) technique~\cite{friedman2008sparse} to serve our purpose. In doing so, we produce a new representation in which SICE serves as the basic unit. An example of this representation is illustrated in Fig.~\ref{fig:SICEch}. 

Exploiting structure sparsity brings the following advantages to this new representation. Firstly, it avoids a direct use of covariance estimate that could be unreliable in the case of small sample, and is completely free of the singularity issue in such case. Secondly, it could be more effective to characterise high-dimensional visual features that usually present structure sparsity, compared with Cov-RP. In addition, the off-diagonal elements in inverse covariance matrix correspond to the partial correlations between two feature components, which factors out the influence of other components, and thus has an advantage over covariance matrix in modeling the essence of data relationship.


Moreover, we extend this new feature representation by utilising the monotonicity property of SICE~\cite{huang2009learning}. Through it, we efficiently obtain a hierarchy of SICE matrices to reflect feature structure at different level of sparsity, and use all of them to produce an enriched representation. Accordingly, two discriminative learning algorithms are developed to adaptively integrate the hierarchy to measure the similarity between samples. This extension not only further improves the recognition performance, but also avoids choosing the single best sparsity level for the SICE process, which could be inefficient and suboptimal in practice. 

To validate our approach and demonstrate its advantages, extensive experimental study is conducted to compare it with existing covariance-based representations and the state-of-the-art comparable methods on various data sets of skeleton-based human action recognition. As will be shown, the proposed feature representation achieves significant improvement over these methods. Especially, compared with the recent methods employing nonlinear kernel techniques, our new representation, as a fully linear technique, shows comparable or even better recognition performance, well demonstrating its potential power. In addition, those  nonlinear kernel methods require prior knowledge to select appropriate kernel functions for the representation, which is not needed in ours.

Our contributions are summarised as follows. (i) To the best of our knowledge, we are the first to improve Cov-RP from the perspective of feature structure modeling and exploit sparsity for this representation. (ii) Our approach produces a new representation based on the SICE matrix, and achieves significant improvement over existing Cov-RP and the other comparable methods. (iii) We extend this new representation to use a hierarchy of SICE matrices and develop discriminative learning algorithms to adaptively integrate them, further improving the recognition performance. (iv) Extensive experimental study is conducted to verify the proposed approach in skeletal human action recognition. In addition, we also demonstrate its remarkable performance in brain image analysis, which shows its potential for generalization. It is worth noting that the proposed method is fundamentally different from that in \cite{zhou2014discriminative}. In that work, SICE was used as a sparse Gaussian model to represent each \textit{class} and each sample was still represented by a feature vector. In contrast, we use SICE as a representation for each \textit{sample}. Accordingly, \cite{zhou2014discriminative} did not classify SICE matrices while we do.

\section{Related Work}
Given a predefined visual feature vector, the statistical variation and mutual correlation of feature components presented on \textit{a set} could be used to represent this set. Cov-RP is based on this idea and implements it effectively. The specific definitions of visual features or the set vary with applications. For example, in the application to object detection at the early days, the features are simply the locations and intensities of each pixel, while the set is an image region (a set of pixels)~\cite{DBLP:conf/cvpr/WangGDD12}. In this case, a small-sized covariance matrix is computed to represent the image region. In the recent application to skeletal human action recognition, the features are the coordinates of skeletal 
joints from a video frame, while the set is a video sequence recording an action instance~\cite{DBLP:conf/ijcai/HusseinTGE13}. Accordingly, a covariance matrix of larger size is obtained to represent this action instance. During the past decade, with its simplicity, robustness with respect to illumination change, and the flexibility of comparing different-sized sets, covariance representations have shown excellent performance in various vision tasks~\cite{DBLP:conf/eccv/TuzelPM06,DBLP:conf/ijcai/HusseinTGE13,DBLP:conf/cvpr/WangGDD12}.    


In the literature, major improvements on Cov-RP generally fall into the following three aspects. The {first} aspect is on computational efficiency. One important improvement is the use of integral image technique to significantly lower the computation on large image regions~\cite{DBLP:conf/eccv/TuzelPM06}. Another recent work uses dimension reduction to reduce the size of covariance matrices while maximally maintaining their  original similarities~\cite{harandi2014manifold}. The {second} aspect focuses on  better evaluating the similarity of covariance matrices. Based on the theories of Riemannian manifold, improvements in this aspect have produced a number of novel measures and algorithms, contributing to the theoretical development of Cov-RP~\cite{quang2014log}. 
 The {third} aspect incorporates nonlinear kernel technique to enhance Cov-RP in modelling complex feature relationship. One way is to nonlinearly map visual features to a kernel-induced feature space and compute covariance matrix therein~\cite{harandi2014bregman}. Another recent way is to compute a kernel matrix whose elements are the kernel values of visual features and use it to replace covariance matrix~\cite{kernelrepiccv}. As shown by that work, this not only largely avoids the singularity issue, but can also model the nonlinear relationship of feature components. 

\section{Proposed new feature representation}
\subsection{Motivation and basic idea}
Our approach starts from reviewing the essence of Cov-RP. It is not difficult to see that this representation is essentially a characterisation of the underlying structure of visual features distributed over a set. In specific, it assumes a Gaussian model and uses the sample-based covariance  estimate to characterise this structure. However, none of the existing methods on Cov-RP has paid sufficient attention to the appropriateness of such covariance estimate. We argue that the following two issues have turned it to be less appropriate. Firstly, the presence of small sample against high feature dimensions makes the sample-based covariance estimate unstable or even singular. It becomes less effective in characterising the underlying data structure. For example, it is well known that the estimates of these larger and smaller eigenvalues intend to be biased in this case, and therefore some kind of regularisation has to be appended. Secondly and importantly, although high-dimensional visual features usually induce complex structure, there is often some prior knowledge available from specific tasks. This valuable prior knowledge shall be adequately incorporated, especially when the sample is scarce. Therefore, rigidly using sample-based covariance estimate is not proper in this case.  

Structure sparsity~\cite{huang2011learning} may be the most common prior knowledge for high-dimensional data. In the terminology of probabilistic graphical model, a distribution can be illustrated as a graph, with each node corresponding to a feature component, and each edge indicating the presence of statistical dependence between the linked two nodes. 
In this case, structure sparsity means the sparsity of the graph, i.e., only a small number of edges exist. A typical example of such situation is in skeletal human action recognition. According to the kinematic configuration of human body, only a small number of joints are \textit{directly} linked. Another more general justification for assuming structure sparsity to high-dimensional data comes from the ``Bet on Sparsity'' principle~\cite{hastie2005elements}. That is, if the graph is truly sparse, we impose a correct prior and will better characterise the underlying structure. If the graph is dense, we will not lose much, because there is no way to recover the underlying structure in the case of small sample. In the literature, sparsity has been well realised in computer vision and implemented in various vision tasks~\cite{huang2011learning}. In this work, we exploit structure sparsity to improve Cov-RP for skeletal human action recognition.  

To impose structure sparsity we switch from covariance matrix to its inverse. This is because covariance matrix measures correlation of feature components, without discriminating direct and indirect correlation. In contrast, inverse covariance measures the partial (direct) correlation by factoring out the effects of other feature components, and this allows the sparsity prior to be conveniently imposed. Note that although inverse covariance has this nice property, it has not been used in Cov-RP before. This may be possibly due to two reasons. Firstly, when covariance matrix is singular, its inverse cannot be readily obtained. Secondly and more importantly, several similarity measures for covariance matrix, such as log-Euclidean kernel~\cite{km2013} and Stein kernel, are inverse invariant. That is, the same result will be obtained if using the inverse directly computed from the original (invertible) covariance matrix. Integrating structure sparsity helps to obtain a more precise inverse covariance, even if the original covariance matrix is less reliable or singular. And the sparse inverse covariance matrix will produce different result from the original covariance matrix. These arguments motivate us to compute sparse inverse covariance estimate to improve Cov-RP.

\subsection{Sparse inverse covariance estimate (SICE)}
Let's denote $\bm{x}$ by a Gaussian model $\mathcal{N}({\boldsymbol\mu}, {\boldsymbol{\mathit\Sigma}})$, where ${\boldsymbol{\mathit\Sigma}}$ denotes covariance matrix and ${\boldsymbol{\mathit\Sigma}}^{-1}$ is its inverse. Each off-diagonal entry of ${\boldsymbol{\mathit\Sigma}}^{-1}$ measures the direct correlation between two feature components. It will be zero if components $i$ and $j$ are conditionally independent given all the remaining ones. 
The estimate of ${\boldsymbol{\mathit\Sigma}}^{-1}$, denoted by $\bm{S}$, can be obtained by maximising a penalised log-likelihood of data, with a symmetric positive definite (SPD) constraint on $\bm{S}$~\cite{friedman2008sparse,huang2009learning}. The optimal solution is called sparse inverse covariance estimate (SICE). 
\begin{equation}\label{eqn:SICE}
\bm{S}^* = \arg \max_{\bm{S} \succ 0}\quad\log \left[\det(\bm{S})\right] - \mathrm{tr}(\hat{\boldsymbol{\mathit\Sigma}}\bm{S}) - \lambda \|\bm{S}\|_1,
\end{equation}where $\hat{\boldsymbol{\mathit\Sigma}}$ is the sample-based covariance matrix, while $\det(\cdot)$, $\mathrm{tr}(\cdot)$ and $\|\cdot\|_1$ denote the determinant, trace and the sum of the absolute values of the entries of a matrix. $||\bm{S}||_1$ imposes sparsity on $\bm{S}$ to achieve more reliable estimation. The tradeoff between the degree of sparsity and the log-likelihood estimation of $\bm{S}$  is controlled by the regularisation parameter $\lambda$. Increasing $\lambda$ value will reveal the underlying data structure at different sparsity levels, with a larger $\lambda$ inducing a sparser $\bm{S}^*$. Note that $\bm{S}^*$ is guaranteed to be SPD and therefore non-singular. 
 In this way, \textit{we obtain a new covariance representation in which sparse inverse covariance estimate is used instead}. 

\subsection{Enriched SICE with hierarchical sparsity}

From the perspective of feature representation, directly using $\bm{S}^*$ may not be ideal due to the existence of the regularisation parameter $\lambda$. It is impractical to tune $\lambda$ for \textit{every} individual sample to obtain a proper feature representation. Even if we use a fixed $\lambda$ value for all samples, this will add one extra algorithmic parameter to the recognition pipeline. Finding the single best $\lambda$ has to resort to multi-fold cross-validation that increases the computation. Another issue is that representing the underlying feature structure at a single sparsity level may not be optimal, as different structures could appear at different levels . The potentially complementary information at other sparsity levels should also be considered.   

To resolve these issues and improve this new representation in further, we propose to 
utilise a nice property of SICE, called ``monotonicity property''~\cite{huang2009learning}. In specific, this property means that by monotonically increasing $\lambda$ in Eq.~(\ref{eqn:SICE}), the resulting  $\bm{S}^*(\lambda)$ will gradually change from being denser to being sparser. The entries of the SICE matrix will gradually vanish and this change is \textit{irreversible}. Therefore, we can use a set of $\lambda$ values arranged as $\lambda_1<\lambda_2<\cdots<\lambda_T$. With this property, we can efficiently and safely obtain a set of $\bm{S}_1^*,\cdots,\bm{S}_T^*$ that guarantee to characterise the underlying feature structure from denser to sparser levels. In doing so, \textit{we obtain an enriched SICE representation, which consists of a hierarchy of SICE matrices}. 
\section{Integration via discriminative learning}
When performing recognition, we need to integrate a hierarchy of SICE matrices to measure the similarity of two samples. The simplest integration may be a convex linear combination. 
Also, it is known that SICE matrices (being SPD) reside on a Riemannian manifold. To respect this fact and make this linear combination more sound, we will perform it in a kernel-induced feature space. The two versions of the combination method, denoted as SICE-RP$_{\boldsymbol{\beta}}$ and  SICE-RP$_{\bm{M}}$, are obtained through two discriminative learning algorithms as follows.
\subsection{SICE-RP$_{\boldsymbol{\beta}}$ method}
A hierarchy of SICE matrices obtained from a sample are mapped from the $d \times d$ dimensional SPD Riemannian manifold into a kernel-induced space $\mathcal{F}$ by a nonlinear mapping $\phi(\cdot): \mathrm{Sym}_{d}^{+} \to \mathcal{F}$. $\phi(\cdot)$ is implicitly conducted by using a kernel (say, log-Euclidean kernel in this paper). 
This $\phi(\cdot)$ mapping brings at least two advantages. Firstly, the Riemannian manifold geometry will be considered by using distance functions specially designed for SPD matrices; Secondly, the images of the SICE matrices under this mapping can be linearly combined in $\mathcal{F}$. 
Specifically, recall that $\mathbb{S} = \{\bm S_1, \bm S_2, \cdots, \bm S_T\}$, $\bm S_j \in \mathrm{Sym}_{d}^{+}$, denoting a hierarchy of SICE matrices extracted from one sample at $T$ different sparsity levels~\footnote{To keep concise, we omit the superscript ``$^{*}$'' from each $\bm S$.}. The linear combination can be expressed as
{\small{\begin{equation}\label{eqn:cominf}
\phi_{\boldsymbol{\beta}}(\mathbb{S}) = \sum_{j = 1}^{T}{\beta_j\phi(\bm S_j)};~~~~\sum_{j = 1}^{T}{\beta_j} = 1;~~~\beta_j \geq 0.
\end{equation}}}
where $\boldsymbol{\beta} = [\beta_1, \beta_2, \cdots, \beta_T]^{'}$ is the combination coefficient. 
We
define a kernel function for the two hierarchies of SICE matrices from samples $p$ and $q$ as
{\small{\begin{equation}\label{eqn:dissackernel}
\begin{split}
k_{\boldsymbol{\beta}}\left(\mathbb{S}^p, \mathbb{S}^q\right) &= \langle \phi_{\boldsymbol{\beta}}(\mathbb{S}^p), \phi_{\boldsymbol{\beta}}(\mathbb{S}^q) \rangle = \sum_{i = 1}^{T}{\sum_{j = 1}^{T}{{\beta}_i \beta}_j \kappa(\bm S_i^p,\bm S_j^q)}= \boldsymbol{\beta}^{'}\bm{\mathit K}(\mathbb{S}^p,\mathbb{S}^q)\boldsymbol{\beta},
\end{split}
\end{equation}}}
where $[\bm{\mathit K}(\mathbb{S}^p,\mathbb{S}^q)]_{ij} = \kappa(\bm S^p_i, \bm S^q_j)$. Note that $k_{\boldsymbol{\beta}}\left(\cdot, \cdot\right)$ and $\kappa\left(\cdot, \cdot\right)$ are two different kernels. The former is defined over two hierarchies of SICE matrices, while the latter is defined over two individual SICE matrices.

\subsection{SICE-RP$_{\bm{M}}$ method}
Different from SICE-RP$_{\boldsymbol{\beta}}$ that assigns a weight to each individual sparsity level, SICE-RP$_{\bm{M}}$ method assigns a weight to each pair of sparsity levels as follows. 
{\small{\begin{equation}
\begin{split}
k_{\bm M}\left(\mathbb{S}^p, \mathbb{S}^q\right) = \langle\bm M,\bm{\mathit K}(\mathbb{S}^p,\mathbb{S}^q)\rangle_{F},
\end{split}
\end{equation}}}
where $\bm M$ is a weight matrix with $M_{ij}$ corresponding to the $(i,j)$th entry of $\bm{\mathit K}(\mathbb{S}^p,\mathbb{S}^q)$. Still imposing the constraint of convex linear combination, $\bm M$ is optimised  by solving:
{\small{\begin{equation}\label{eq:mrmbm}
\begin{aligned}
 \bm M^* = \arg \min_{{\sum_{i,j = 1}^{T}{\bm M}_{ij}} = 1; {\bm M}_{ij} \geq 0}~~~  {R}^2\|\bm{w}\|^2~~.
\end{aligned}
\end{equation}}} It is not difficult to see that the above SICE-RP$_{\boldsymbol{\beta}}$ is a special case of SICE-RP$_{\bm M}$, because $k_{\boldsymbol{\beta}}\left(\mathbb{S}^p, \mathbb{S}^q\right)=\langle\boldsymbol{\beta}^{'}\boldsymbol{\beta},\bm{\mathit K}(\mathbb{S}^p,\mathbb{S}^q)\rangle_{F}$. In SICE-RP$_{\boldsymbol{\beta}}$, $\bm M$ is restricted to a rank-one matrix $\boldsymbol{\beta}^{'}\boldsymbol{\beta}$. Therefore, SICE-RP$_{\bm{M}}$ has more flexibility to weight these cross-sparsity-level kernel evaluations.
\subsection{Optimisation}

The combination coefficient $\boldsymbol{\beta}$ or $\bm{M}$ can then be viewed as the tunable parameter of the kernel $k_{\boldsymbol{\beta}}\left(\cdot, \cdot\right)$ or $k_{\bm{M}}\left(\cdot, \cdot\right)$, and its value can be sought by optimising a generalisation bound on classification performance, e.g., the radius margin bound that is the upper bound of Leave-One-Out error~\cite{chapelle2002choosing}. 
 In the following, the optimisation of $\boldsymbol{\beta}$ is given and $\bm{M}$ can be obtained similarly.


We first consider a binary classification task and then extend the result to the multi-class case.
Given a training set of $N$ samples, and without loss of generality, the samples are labeled by $l \in\{-1, 1\}$, the optimal $\boldsymbol{\beta}$ can be obtained by solving
{\small{
\begin{equation}\label{eq:mrmbtheta}
\begin{aligned}
 \boldsymbol{\beta}^* = \arg \min_{\sum_{j = 1}^{T}{\beta_j} = 1; \beta_j \geq 0}~~~  {R}^2\|\bm{w}\|^2~~.
\end{aligned}
\end{equation}}}
where $R$ is the radius of the smallest sphere enclosing all the training samples, while $\bm w$ denotes the normal of the SVM separating hyperplane, with $1/\|\bm w\|$ being the margin. ${R}^2$ can be obtained by optimising the following problem:
{\small{
\begin{equation}\label{eq:r2}
\begin{aligned}
{R}^2 &= \max_{\alpha \in \mathbb{R}^N} \Big [ \sum_{i = 1}^{N} \alpha_i k\left(\mathbb{S}^i, \mathbb{S}^i\right) -  \sum_{i,j = 1}^{N}\alpha_i \alpha_j k\left(\mathbb{S}^i, \mathbb{S}^j\right) \Big]\\
& \text{subject to:}   ~\sum_{i = 1}^{N} \alpha_i = 1;~~ \alpha_i\geq 0 ~~(i = 1, 2, \dots, l).
\end{aligned}
\end{equation}}}
where $k\left(\mathbb{S}^i, \mathbb{S}^j\right)$ denotes the kernel function $k_{\boldsymbol{\beta}}\left(\cdot, \cdot\right)$ or $k_{\bm M}\left(\cdot, \cdot\right)$ defined over two hierarchies of SICE matrices $\mathbb{S}^i$ and $\mathbb{S}^j$ in the paper. And $\|\bm{w}\|^2$ can be obtained by solving the following optimisation problem of SVM with $L_2$-norm soft margin:
{\small{
\begin{equation}\label{eq:svm}
\begin{aligned}
&\frac{1}{2} \| \bm{w}\|^2 = &&\max_{\boldsymbol{\eta} \in \mathbb{R}^N}  \Big [ \sum_{i = 1}^{N} \eta_i - \frac{1}{2} \sum_{i,j = 1}^{N}\eta_i \eta_j l_{i} l_{j} \tilde{k}\left(\mathbb{S}^i, \mathbb{S}^j\right) \Big]\\
& \text{subject to:} &&  \sum_{i = 1}^{N} \eta_i l_{i} = 0;~ \eta_i\geq 0 ~(i = 1, 2, \dots, N)
\end{aligned}
\end{equation}}}
where $\tilde{k}\left(\mathbb{S}^i, \mathbb{S}^j\right)$ $= {k}\left(\mathbb{S}^i, \mathbb{S}^j\right) + \frac{1}{C}\delta_{ij}$;  
 $C$ is the regularisation parameter; $\delta_{ij} = 1$ if $i = j$, and $0$ otherwise.

How to solve Eq.(\ref{eq:mrmbtheta}) has been well studied in the literature. In brief, it can be optimised by iteratively 1) updating $R^2$ and $\|\bm{w}\|^2$; 2) minimizing $R^2\|\bm{w}\|^2$  with respect to $\boldsymbol{\beta}$ using gradient-based methods,
as outlined in Algorithm \ref{alg:1}.

\begin{algorithm}[!tb] %
\renewcommand{\algorithmicrequire}{\textbf{Input:}}
\renewcommand\algorithmicensure {\textbf{Output:} }
\caption{Proposed SICE-RP$_{\boldsymbol{\beta}}$ method with the radius margin bound.} %
\label{alg:1} %
\begin{algorithmic}[1] %
\REQUIRE 
A training set $\{(\mathbb{S}_i,l_{i})\}_{i=1}^{N}$, stopping criteria: i) The total number of iterations $I = 100$; ii) A small positive value $\tau = 10^{-5}$.
\ENSURE 
${\boldsymbol{\beta}}$.~~\\
\FOR { $i =  1:I$}
    \STATE Solve $R^2$ and $\|\bm{w}\|^2$ according to Eq. (\ref{eq:r2}) and Eq. (\ref{eq:svm});
    \STATE Update ${\boldsymbol{\beta}}$ by a gradient-based method;
    \IF{$|J_{i+1} - J_{i}| \leq \tau J_{i}$~~($J$ is defined as $R^2\|\bm{w}\|^2$)}
     \STATE Break;
     \ENDIF
\ENDFOR
\RETURN ${\boldsymbol{\beta}}$; %
\end{algorithmic}
\end{algorithm}
For multi-class classification tasks, we employ one-vs-one partitioning strategy and optimize $\boldsymbol{\beta}$ by using a pairwise combination of the radius margin bounds of binary SVM classifiers. Please refer to the supplementary for more details.
%


\subsection{Differences from MKL and EMK}\label{CSSIvsMKLEMK}
Multiple kernel learning (MKL) has been commonly used to combine different sources of information. Also, efficient match kernel (EMK)~\cite{bo2009efficient} has been well-known for evaluating the similarity of two sets of points.  With our notations, they can be expressed as follows, respectively. 
{\small{
\begin{displaymath}
k_{\mathrm{MKL}}\left(\mathbb{S}^p, \mathbb{S}^q\right)=\sum_{j = 1}^{T}{\beta_j}\kappa(\bm S_j^p,\bm S_j^q),~~~~~~~k_{\mathrm{EMK}}\left(\mathbb{S}^p, \mathbb{S}^q\right)=\frac{1}{T^2}\sum_{i = 1}^{T}{\sum_{j = 1}^{T}\kappa(\bm S_i^p,\bm S_j^q)}.
\end{displaymath}}}
Comparing them with our methods shows that i) MKL and the proposed integration methods differ in cross-sparsity-level comparison. MKL is often used to integrate heterogeneous sources, and different sources are usually not comparable. As a result, MKL only considers the similarity between samples from the \textit{same} source, i.e. $\bm S_j^p$ and $\bm S_j^q$, $j \in \lbrack1,T\rbrack$. In our case, SICE matrices at different sparsity levels are of the same type and therefore comparable. This allows us to explore the similarity across sources, i.e. $S_i^p$ and $S_j^q$, $i,j \in \lbrack1, T\rbrack$ to measure the similarity between two samples. In this sense, our method enjoys more flexibility to capture information than MKL.
%
ii) EMK uses an equal weight to combine the similarity measure over different pairs. In contrast, the weights in our methods are adaptively learned, allowing us to better align with a task. 


\section{Computational issues}
Given a $d\times{d}$ sample-based covariance matrix $\hat{\boldsymbol{\mathit\Sigma}}$ estimated from $m$ feature vectors, the optimisation in Eq.~(\ref{eqn:SICE}) is proved to be convex and guaranteed to converge even when $m < d$~\cite{friedman2008sparse} and  SICE matrix can be efficiently obtained by the off-the-shelf package GLASSO~\cite{friedman2008sparse} in $\mathcal O(d^3)$. As shown in~\cite{friedman2008sparse}, it takes only $0.014$ CPU second to obtain a $100 \times 100$ SICE matrix. It also allows to efficiently build a path of SICE matrices for different values of $\lambda$. Therefore, the proposed methods can be efficiently computed. In addition, note that the complexity of SICE-RP is independent of the feature number $m$ once $\hat{\boldsymbol{\mathit\Sigma}}$ is provided. Also, as Cov-RP, SICE-RP still allows two sets with different number of features to be compared, because the resulting SICE matrix has a fixed size of $d \times d$. 
\section{Experimental result}
We compare our three proposed methods (i.e., SICE-RP, SICE-RP$_{\boldsymbol{\beta}}$ and SICE-RP$_{\bm{M}}$)  with both the classic Cov-RP and several state-of-the-art methods mainly in skeletal human action recognition. Four benchmark data sets are tested, including HDM05~\cite{harandi2014bregman}, MSRC-12~\cite{DBLP:conf/ijcai/HusseinTGE13}, MSR-Action3D~\cite{luo2013group} and MSR-DailyActivity3D\\~\cite{yang2014super}. For all data sets, only \textit{skeleton} data are used while other information (e.g., depth maps or RGB videos) is not utilised.  The proposed methods are also evaluated in medical image analysis to demonstrate their potential for generalization.   


A  kernel SVM classifier is employed throughout all experiments, in which the log-Euclidean kernel~\cite{km2013} is used to measure the similarity of two SPD matrices. For a fair comparison, all algorithmic parameters
are tuned by multi-fold cross-validation on the training set only. The sparsity parameter $\lambda$ used in SICE-RP is also chosen by cross-validation on the training set. For the proposed integration methods, ten sparsity levels of SICEs  
 are computed for each sample,  corresponding to the very dense to very sparse representations. To compare with the state-of-the-art methods, the training and test sets of these data sets are partitioned by following the literature. For HDM05, we used the instances of two subjects for training and the remaining for test, as in~\cite{harandi2014manifold}. For MSRC-12, MSR-Action3D and MSR-DailyActivity3D, the cross-subject test setting~\cite{li2010action} is used, i.e., the odd-indexed subjects for training and the even-indexed ones  for test. 

Features are generated as follows. For HDM05 and MSRC-12, the 3D coordinates of each joint are used as the frame features, leading to a feature dimensionality of $93$~(${3}\times{31}$ joints) in HDM05 and $60$~(${3}\times{20}$ joints) in MSRC-12. For MSR-Action3D and MSR-DailyActivity3D, velocity is used as the frame features~\cite{Zanfir2013}, which is calculated by the coordinate difference of 3D skeleton joints between a frame and its two direct neighbor frames. The dimensionality of the frame feature is $120~(2\times{3}\times{20}$ joints). The frame number in each action instance is $30\sim700$ in HDM05, $50\sim300$ in MSRC-12, $40\sim60$ in MSR-Action3D and $120\sim500$ in MSR-DailyActivity3D. In Cov-RP, to address the singularity issue,  a small regulariser $\lambda{\bm I}$ (e.g., $\lambda=10^{-7}$) is appended as in~\cite{DBLP:conf/cvpr/WangGDD12}.
\subsubsection{Result on HDM05 data set}
HDM05 has about $1500$ instances from over $100$ motion classes. Most classes have $10$ to $50$ realisations of five actors named ``bd'', ``bk'', ``dg'', ``mm'' and ``tr''. We use two subjects ``bd'' and ``mm'' for training and the remaining three for test by following~\cite{harandi2014manifold}. 
The results are given in Table~\ref{tab:HDM0514}.

In addition to Cov-RP, the results of another six methods in the literature are also quoted in Table~\ref{tab:HDM0514}. These methods can be roughly categorised into linear and nonlinear representations (corresponding to the lower or the upper portion of Table~\ref{tab:HDM0514}, respectively). As for the linear representation methods, RSR~\cite{sc2012}, RSR-ML~\cite{harandi2014manifold} and CDL~\cite{DBLP:conf/cvpr/WangGDD12} are covariance-based representations, but they further conduct dimensionality reduction on covariance matrix using sparse coding or projection. We also test InverseCov-RP, which  directly uses the inverse of the covariance matrix as representation without exploiting structure sparsity. As for the nonlinear representation methods, Cov-{$J_{\mathcal{H}}$-SVM~\cite{harandi2014bregman} employs an infinite-dimensional covariance matrix in a kernel-induced feature space as representation. Ker-RP-RBF and Ker-RP-POL~\cite{kernelrepiccv} are the two variants of the recently proposed non-linear kernel representations using RBF and polynomial kernels~\cite{kernelrepiccv}. Note that, our methods
belong to the linear representations.

 We follow~\cite{harandi2014manifold} and test the classification accuracy on i) $14$ classes only (the left column in  Table~\ref{tab:HDM0514}), 
and ii) all the 100 classes (the right column in  Table~\ref{tab:HDM0514}). 

For $14$ classes, Cov-RP shows quite competitive performance and outperforms the linear representations of RSR, RSR-ML, CDL and the nonlinear representation Cov-{$J_{\mathcal{H}}$-SVM.  As expected, InverseCov-RP obtains the same performance as Cov-RP since the log-Euclidean kernel is inverse-invariant. The nonlinear kernel representations of Ker-RP-RBF and Ker-RP-POL~\cite{kernelrepiccv} outperform Cov-RP. In comparison, the three proposed methods demonstrate remarkable performance. SICE-RP achieves a high classification accuracy of $96.8$\%,  on a par with 
 Ker-RP-RBF~\cite{kernelrepiccv} and better than all the other quoted methods. This may indicate the efficacy of exploring the structure sparsity. Moreover,  
SICE-RP$_{\bm{M}}$ further boosts the classification accuracy from $96.8$\% to $97.3$\%, updating the state-of-the-art performance.

For all the 100 classes, the overall classification accuracy decreases due to the significant increase of the number of action classes. In this case, the proposed SICE-RP still outperforms all the quoted ones in comparison. It achieves a significant improvement of $8.7$ percentage points over Cov-RP, and even wins the non-linear kernel representation methods Ker-RP-RBF and Ker-RP-POL~\cite{kernelrepiccv}. When integrating the hierarchy of SICEs by the proposed methods, the improvement becomes more salient. Specifically,  SICE-RP$_{\boldsymbol{\beta}}$ achieves a classification accuracy of $69.3$\%, which is $10.4$ percentage points higher than Cov-RP and $3.1$ percentage points higher than Ker-RP-RBF~\cite{kernelrepiccv}.

\begin{table}[!htbp]
\centering
\begin{minipage}[t]{0.48\textwidth}
\centering
\scriptsize
\caption{Comparison on HDM05 data set (Two experiments).}
\label{tab:HDM0514} \centering
\begin{tabular}{l|cc}
					 & $14$ classes & All classes \\
Methods in comparison & Accuracy & Accuracy \\
\hline
\multicolumn{3}{c}{Methods using nonlinear representation}\\
\hline
{Cov-$J_{\mathcal{H}}$-SVM~\cite{harandi2014bregman}}&$82.5$ & Not reported\\
Ker-RP-POL~\cite{kernelrepiccv} & $93.6$ & $64.3$\\
Ker-RP-RBF~\cite{kernelrepiccv} &${96.8}$ & ${66.2}$\\
\hline
\multicolumn{3}{c}{Methods using linear representation}\\
\hline
{RSR~\cite{sc2012}}&$76.1$ & Not reported\\
{RSR-ML~\cite{harandi2014manifold}}&$81.9$ & $40.0$\\
{CDL~\cite{DBLP:conf/cvpr/WangGDD12}}&$79.8$ & Not reported\\
Cov-RP~\cite{DBLP:conf/eccv/TuzelPM06} &$91.5$&$58.9$\\
InverseCov-RP &$91.5$&$58.9$\\
SICE-RP (proposed)&$96.8$&$67.6$\\
SICE-RP$_{\boldsymbol{\beta}}$ (proposed)&$96.8$&$\bm{69.3}$\\
SICE-RP$_{\bm{M}}$ (proposed)&$\bm{97.3}$&${69.1}$\\
\end{tabular}
\end{minipage}
\hspace{3mm}
\begin{minipage}[t]{0.46\textwidth}
\centering
\scriptsize
\caption{Comparison on MSRC-12 data set.}
\label{tab:MSRC-12} \centering
\begin{tabular}{l|c}
Methods in comparison & Accuracy\\
\hline
\multicolumn{2}{c}{Methods using nonlinear representation}\\
\hline
{Cov-$J_{\mathcal{H}}$-SVM~\cite{harandi2014bregman}}&89.8\\
Ker-RP-POL~\cite{kernelrepiccv} &$90.5$\\
Ker-RP-RBF~\cite{kernelrepiccv} &${92.3}$\\
\hline
\multicolumn{2}{c}{Methods using linear representation}\\
\hline
{Hierarchy of Cov3DJs~\cite{DBLP:conf/ijcai/HusseinTGE13}}&$91.7$\\
Cov-RP~\cite{DBLP:conf/eccv/TuzelPM06} &$89.2$\\
InverseCov-RP&$89.2$\\
SICE-RP (proposed)&$92.5$\\
SICE-RP$_{\boldsymbol{\beta}}$ (proposed)&$92.8$\\
SICE-RP$_{\bm{M}}$ (proposed)&$\bm{93.3}$\\
\end{tabular}
\end{minipage}
\end{table}



\subsubsection{Result on MSRC-12 data set}
MSRC-12 contains $12$ gesture categories from $30$ subjects. 
As shown in Table~\ref{tab:MSRC-12}, SICE-RP again outperforms all the existing methods, including Cov-RP and the non-linear kernel representation methods in~\cite{kernelrepiccv}. 
By integrating the hierarchy of SICEs via our proposed integration methods, the classification accuracy of SICE-RP can be further improved to $93.3$\% by SICE-RP$_{\bm{M}}$ and to $92.8$\% by SICE-RP$_{\boldsymbol{\beta}}$. This reinforces the effectiveness of the proposed SICE-RP and the integration methods. 

In addition, to provide insight on the proposed representation, we visualise the SICE matrices computed on this data set to show the identified underlying structure of the interactions between skeletal joints for different actions. The results can be found in Fig.~\ref{fig:SICEch} and the supplementary material. 



\subsubsection{Result on MSR-Action3D data set}
MSR-Action3D contains $20$ categories of actions from ten subjects. 
Each action is performed two or three times by each subject. 
The results are given in Table~\ref{tab:MSR-Action3D}. Note that several non-Cov-related methods in the literature are also quoted in the left portion of this table.

As can be seen, although Cov-RP performs poorly in this case, 
the proposed SICE-RP achieves an accuracy up to  $96.5$\%, bringing an improvement of $20$ percentage points over Cov-RP, and $16$ percentage points over  Cov-{$J_{\mathcal{H}}$-SVM. 
It is interesting to see that SICE-RP also wins the methods in the left portion of Table~\ref{tab:MSR-Action3D}, which involve complex representations of features, e.g., sparse coding~\cite{yang2014super} or use additional information like depth maps~\cite{zhu2013fusing}. When a hierarchy of SICEs are integrated by the proposed SICE-RP$_{\boldsymbol{\beta}}$ and SICE-RP$_{\bm{M}}$, the performance can be further boosted to $96.9$\%, reaching the state-of-the-art performance of the nonlinear representation Ker-RP-RBF~\cite{kernelrepiccv}. 
Although SICE-RP$_{\boldsymbol{\beta}}$ and SICE-RP$_{\bm{M}}$ tie  Ker-RP-RBF in~\cite{kernelrepiccv},  they have an advantage: the methods in~\cite{kernelrepiccv} require prior knowledge to select appropriate kernel functions for the representation, which is not needed in SICE-RP$_{\boldsymbol{\beta}}$ or SICE-RP$_{\bm{M}}$.

\begin{table}[!ht]
\scriptsize
\caption{Comparison on MSR-Action3D data set.}
\label{tab:MSR-Action3D} \centering
\begin{tabular}{l|c|l|c}
Methods in comparison & Accuracy&Methods using nonlinear representation&Accuracy\\
\hline
{Structured Streaming Skeletons~\cite{zhao2013online}}&$81.7$&Cov-{$J_{\mathcal{H}}$-SVM~\cite{harandi2014bregman}}&$80.4$\\
{DBN+HMM~\cite{wu2014leveraging}}&$82.0$&Ker-RP-POL~\cite{kernelrepiccv} &$96.2$\\
{Actionlet Ensemble~ \cite{wang2014learning}}&$88.0$&Ker-RP-RBF~\cite{kernelrepiccv} & ${\bm{96.9}}$\\
\cline{3-4}
{Pose Set~\cite{wang2013approach}}&$90.0$&Methods using linear representation&\\
\cline{3-4}
{Moving Pose~ \cite{Zanfir2013}}&$91.7$&{Hierarchy of Cov3DJs~\cite{DBLP:conf/ijcai/HusseinTGE13}}&$90.5$\\
{Lie Group~\cite{vemulapalli2014human}}&$92.5$&Cov-RP~\cite{DBLP:conf/eccv/TuzelPM06} &$74.0$\\
{SNV~ \cite{yang2014super}}&$93.1$&InverseCov-RP &$74.0$\\
{Spatiotemp. Features Fusing~\cite{zhu2013fusing}}&$94.3$&SICE-RP (proposed)&$96.5$\\
{DL-GSGC+TPM~\cite{luo2013group}}&$96.7$&SICE-RP$_{\boldsymbol{\beta}}$ (proposed)&$\bm{96.9}$\\
&&SICE-RP$_{\bm{M}}$ (proposed)&$\bm{96.9}$\\
\hline
\end{tabular}
\end{table}

\subsubsection{Result on MSR-DailyActivity3D data set}
MSR-DailyActivity3D involves human-object interactions 
such as \textit{drink}, \textit{eat}, \textit{read book}, etc. 
The results are given in Table~\ref{tab:MSR-DailyActivity3D}. 
On this data set, the best performance is achieved by the non-linear representation methods Ker-RP-POL and Ker-RP-RBF~\cite{kernelrepiccv}.  Cov-RP performs close to some of the state-of-the-art  non-Cov-based representations. Our SICE-RP once again demonstrates reasonably good performance, with an accuracy of $93.1$\%,  significantly better than most of the  quoted state-of-the-art results. Specifically, it outperforms Cov-RP by a large margin of $8.1$ percentage points. The accuracy can be further improved to  $95.0$\% through integrating multiple SICEs using SICE-RP$_{\bm{M}}$ or SICE-RP$_{\boldsymbol{\beta}}$. This result is close to the highest accuracy of  $96.9$\% obtained by Ker-RP-POL.  Note that, additional information is used in some state-of-the-art methods, such as depth map~\cite{oreifej2013hon4d,yang2014super} and local occupancy patterns~\cite{wang2014learning}, while our SICE-RP soley utilizes the skeleton data. 

\begin{table}[htbp]
\centering
\begin{minipage}[t]{0.46\textwidth}
\centering
\scriptsize
\caption{Comparison on MSR-DailyActivity3D data set.}
\label{tab:MSR-DailyActivity3D} \centering
\begin{tabular}{l|c}
Methods in comparison & Accuracy\\
\hline
{Moving Pose~\cite{Zanfir2013}}&$73.8$\\
{Local HON4D~\cite{oreifej2013hon4d}}&$80.0$\\
{Actionlet Ensemble~\cite{wang2014learning}}&$86.0$\\
{SNV~\cite{yang2014super}}&$86.3$\\
\hline
\multicolumn{2}{c}{Methods using nonlinear representation}\\
\hline
{Cov-$J_{\mathcal{H}}$-SVM~\cite{harandi2014bregman}}&$75.0$\\
Ker-RP-POL~\cite{kernelrepiccv} &$\bm{96.9}$\\
Ker-RP-RBF~\cite{kernelrepiccv} &{$96.3$}\\
\hline
\multicolumn{2}{c}{Methods using linear representation}\\
\hline
Cov-RP~\cite{DBLP:conf/eccv/TuzelPM06} &$85.0$\\
InverseCov-RP &$85.0$\\
SICE-RP (proposed)&$93.1$\\
SICE-RP$_{\boldsymbol{\beta}}$ (proposed)&$95.0$\\
SICE-RP$_{\bm{M}}$ (proposed)&${95.0}$\\
\end{tabular}
\end{minipage}
\hspace{3mm}
\begin{minipage}[t]{0.47\textwidth}
\centering
\scriptsize
\caption{Comparison on ADHD-200 data set.}
\label{tab:ADHD} \centering
\begin{tabular}{l|c}
Methods in comparison & Accuracy\\
\hline
AttributedGraph\cite{dey2014attributed}&62.8\\
NetStructure\cite{dey2012exploiting}&69.6\\
\hline
\multicolumn{2}{c}{Methods using nonlinear representation}\\
\hline
{Cov-$J_{\mathcal{H}}$-SVM~\cite{harandi2014bregman}}&67.8\\
Ker-RP-POL~\cite{kernelrepiccv} &$54.3$\\
Ker-RP-RBF~\cite{kernelrepiccv} &${58.4}$\\
\hline
\multicolumn{2}{c}{Methods using linear representation}\\
\hline
Cov-RP~\cite{DBLP:conf/eccv/TuzelPM06} &$55.8$\\
InverseCov-RP&$55.8$\\
SICE-RP (proposed)&$69.0$\\
SICE-RP$_{\boldsymbol{\beta}}$ (proposed)&$72.5$\\
SICE-RP$_{\bm{M}}$ (proposed)&$\bm{73.2}$\\
\end{tabular}
\end{minipage}
\end{table}

\subsubsection{Comparison on medical image analysis}

Cov-RP is also used in brain image analysis. The benchmark data set  ADHD-200 is tested for this case to verify the potential generalisation of the proposed methods. It is provided by the Neuro Bureau for differentiating Attention Deficit Hyperactivity Disorder (ADHD) from healthy control subjects.  
ADHD-200 consists of resting-state functional MRI (fMRI) images of $768$ training and $197$ test subjects.
The fMRI images are preprocessed with Athena pipeline\footnote{\url{http://neurobureau.projects.nitrc.org/ADHD200/Introduction.html}}, after which, each subject is characterised by the averaged time series from each of $90$ brain regions. A $90\times{90}$ covariance matrix is estimated for each subject based on $70 \sim 100$ time points.Therefore this task suffers the issue of small sample vs high feature dimensionality, similar to the skeletal human action recognition task. 

In the literature, the state-of-the-art classification accuracy on this data set is $69.6$\%  in~\cite{dey2012exploiting}. The comparison between the existing methods and ours is presented in Table~\ref{tab:ADHD}. As seen, Cov-RP (of brain regions) only obtains an accuracy of $55.8$\%, much worse than that in~\cite{dey2012exploiting}. This is probably due to the unreliable covariance estimation suffering from the small sample problem. The nonlinear representation Ker-RP-RBF~\cite{kernelrepiccv} also performs poorly with a classification accuracy below $60.0$\%. On the contrary, SICE-RP obtains an accuracy of $69.0$\%, beating both Cov-RP and Ker-RP-RBF and comparable with that in~\cite{dey2012exploiting}. This may be attributed to the fact that brain network is very sparse. The integration of SICEs seems very effective on this data set, as SICE-RP$_{\boldsymbol{\beta}}$ and SICE-RP$_{\bm{M}}$ significantly boost the classification accuracy to $72.5$\% and $73.2$\%, respectively. This demonstrates that our methods could be potentially generalised to  other applications with small sample and high dimensionality.

We have verified the effectiveness of exploiting structure sparsity in skeletal human action recognition and medical image analysis. For these tasks, the number of samples is relatively small and the dimensionality is high. Also, the prior knowledge on structure sparsity is clear in these tasks. As a sanity check, we further investigate how the proposed methods perform on the tasks with lower feature dimensions and a larger number of feature vectors. This sanity check experiment agrees with the principle of ``Bet on sparsity'' and suggests that exploiting structure sparsity could well maintain competitive performance and be a safe option for more applications. Refer to the supplementary for details.

\subsubsection{Comparison with MKL and EMK}
As shown in section~\ref{CSSIvsMKLEMK}, MKL does not utilise the cross-source information and is a special case of  SICE-RP$_{\bm{M}}$. An experiment is conducted to compare SICE-RP$_{\boldsymbol{\beta}}$, SICE-RP$_{\bm{M}}$ and MKL to investigate if the cross-source information helps. The four human action recognition data sets are used. As seen in Table~\ref{tab:CSSIVSMKL}, MKL is only able to improve the performance of SICE-RP on MSRC-12 and MSR-DailyActivity3D. However, this is still inferior to the best classification performance achieved by our proposed  SICE-RP$_{\boldsymbol{\beta}}$ and SICE-RP$_{\bm{M}}$. This experiment also compares SICE-RP$_{\boldsymbol{\beta}}$, SICE-RP$_{\bm{M}}$ and EMK. 
As seen, EMK only improves SICE-RP on MSRC-12 and HDM05 (100 classes), but is worse than SICE-RP$_{\boldsymbol{\beta}}$ and SICE-RP$_{\bm{M}}$ on all action recognition data sets. This demonstrates the advantage of our adaptive integration methods.
\begin{table}[!ht]
\scriptsize
\caption{Comparison between SICE-RP$_{\boldsymbol{\beta}}$, SICE-RP$_{\bm{M}}$, MKL and EMK on human action recognition data sets.}
\label{tab:CSSIVSMKL} \centering
\begin{tabular}{p{100 pt}|p{50 pt}p{50 pt}p{50 pt}p{30 pt}p{30 pt}}
Data set & SICE-RP& SICE-RP$_{\boldsymbol{\beta}}$ &SICE-RP$_{\bm{M}}$& MKL & EMK\\
\hline
MSRC-12&$92.5$&$92.8$&$\bm{93.3}$&$93.1$&$92.7$\\
HDM05 (14 classes)&$96.8$&$96.8$&$\bm{97.3}$&$96.8$&$96.3$\\
HDM05 (100 classes)&$67.6$&$\bm{69.3}$&$69.1$&$66.5$&$68.6$\\
MSR-Action3D&$96.5$&$\bm{96.9}$&$\bm{96.9}$&$95.8$&$95.7$\\
MSR-DailyActivity3D&$93.1$&$\bm{95.0}$&$\bm{95.0}$&$94.4$&$92.5$\\
\hline
Average&$89.3$&$\bm{90.2}$&$\bm{90.3}$&$89.3$&$89.2$\\
\hline
\end{tabular}
\end{table}

\section{Conclusion and future work}
To address the new issues encountered by covariance representation, we propose to improve the quality of characterising the underlying structure of visual features, and this leads to the use of SICE matrix as a generic feature representation. This new representation exploits the structure sparsity potentially existing among feature components, and is therefore more robust against sample scarcity and high feature dimensionality. The significant improvement achieved by this new representation is verified in skeletal human action recognition and medical image analysis. Also, the two integration methods developed in this work further improve recognition performance while avoiding searching for a single best sparsity level. The future work will apply the proposed representation to more vision tasks, investigate its efficacy for unsupervised learning scenario, and explore its interaction with nonlinear representations.  
{\small
\bibliographystyle{splncs}
\bibliography{iccv15_short,tnn}
}

\end{document}